%% file: 1022.tex
\documentclass[10pt,twocolumn,letterpaper]{article}

\usepackage{iccv}
\usepackage{times}
\usepackage{epsfig}
\usepackage{graphicx}
\usepackage{amsmath}
\usepackage{amssymb}

\usepackage[utf8]{inputenc} 
\usepackage[T1]{fontenc}    
\usepackage{booktabs}       
\usepackage{amsfonts}       
\usepackage{nicefrac}       
\usepackage{microtype}      
\usepackage{color}
\usepackage{colortbl}
\usepackage{tabularx}
\usepackage{arydshln}
\usepackage{array,multirow}
\usepackage{xfrac}
\usepackage{bm}
\usepackage{soul}
\usepackage[export]{adjustbox}
\usepackage{graphicx}
\usepackage{verbatim}
\usepackage{caption}
\usepackage{subcaption}
\usepackage{mwe}
\usepackage[shortcuts]{extdash}
\usepackage[toc,page]{appendix}
\usepackage[ruled,lined,linesnumberedhidden]{algorithm2e}
\usepackage{enumitem}

\usepackage[pagebackref=true,breaklinks=true,letterpaper=true,colorlinks,bookmarks=false]{hyperref}

\newcolumntype{C}[1]{>{\centering\let\newline\\\arraybackslash\hspace{0pt}}m{#1}}
\newcolumntype{K}[1]{>{\centering\arraybackslash}p{#1}}

\iccvfinalcopy 

\def\iccvPaperID{1022} 

\ificcvfinal\fi
\begin{document}

\title{MarioQA: Answering Questions by Watching Gameplay Videos}

\author{Jonghwan Mun\thanks{Both authors contributed equally.} \quad\quad\quad Paul Hongsuck Seo\footnotemark[1] \quad\quad\quad Ilchae Jung \quad\quad\quad Bohyung Han\\
Department of Computer Science and Engineering, POSTECH, Korea\\
{\tt\small \{choco1916, hsseo, chey0313, bhhan\}@postech.ac.kr}
}

\maketitle
\thispagestyle{empty}

\input{abstract.tex}
\input{introduction.tex}
\input{related_work.tex}
\input{dataset.tex}

\input{spatiotemporal.tex}
\input{experiments.tex}
\input{conclusion.tex}


\paragraph{Acknowledgement}
This work was partly supported by the ICT R\&D program of MSIP/IITP [2014-0-00059 and 2016-0-00563] and the NRF grant [NRF-2011-0031648] in Korea.
Samsung funded our preliminary work in part.

{\small
\bibliographystyle{ieee}
\bibliography{egbib}
}

\end{document}

%% file: abstract.tex

\begin{abstract}
We present a framework to analyze various aspects of models for video question answering (VideoQA) using customizable synthetic datasets, which are constructed automatically from gameplay videos.
Our work is motivated by the fact that existing models are often tested only on datasets that require excessively high-level reasoning or mostly contain instances accessible through single frame inferences.
Hence, it is difficult to measure capacity and flexibility of trained models, and existing techniques often rely on ad-hoc implementations of deep neural networks without clear insight into datasets and models.
We are particularly interested in understanding temporal relationships between video events to solve VideoQA problems; this is because reasoning temporal dependency is one of the most distinct components in videos from images.
To address this objective, we automatically generate a customized synthetic VideoQA dataset using {\em Super Mario Bros.} gameplay videos so that it contains events with different levels of reasoning complexity.
Using the dataset, we show that properly constructed datasets with events in various complexity levels are critical to learn effective models and improve overall performance.
\end{abstract}

%% file: introduction.tex

\section{Introduction}

While deep convolutional neural networks trained on large-scale datasets have been making significant progress on various visual recognition problems, most of these tasks focus on the recognition in the same or similar levels, \eg, objects~\cite{Vgg16,Googlenet}, scenes~\cite{Scenerecognition}, actions~\cite{twoStream,C3D}, attributes~\cite{Attribute,FaceAttribute}, face identities~\cite{Deepface}, etc.
On the other hand, image question answering (ImageQA)~\cite{composeVQA,VQA,abcAtt,MCBPooling,DPPNet,whereAtt,askMe,dynamicMem,askAtt,stackedAtt,simpleVQA} addresses a holistic image understanding problem, and handles diverse recognition tasks in a single framework.
The main objective of ImageQA is to find an answer relevant to a pair of an input image and a question by capturing information in various semantic levels.
This problem is often formulated with deep neural networks, and has been successfully investigated thanks to advance of representation learning techniques and release of outstanding pretrained deep neural network models~\cite{composeVQA,MCBPooling,DPPNet,dynamicMem,stackedAtt}.

Video question answering (VideoQA) is a more challenging task and is recently introduced as a natural extension of ImageQA in~\cite{LSMDCFIB,MovieQA,VideoQA}.
In VideoQA, it is possible to ask a wide range of questions about temporal relationship between events such as dynamics, sequence, and causality.
However, there is only limited understanding about how effective VideoQA models in capturing various information from videos, which is partly because there is no proper framework including dataset to analyze models.
In other words, it is not straightforward to identify main reason of failure in VideoQA problems---dataset vs. trained model.

There exist a few independent datasets for VideoQA~\cite{LSMDCFIB,MovieQA,VideoQA}, but they are not well-organized enough to estimate capacity and flexibility of models accurately.
For example, answering questions in MovieQA dataset~\cite{MovieQA} requires too high-level understanding about movie contents, which is almost impossible to extract from visual cues only (\eg, calling off one's tour, corrupt business, and vulnerable people) and consequently needs external information or additional modalities.
On the other hand, questions in other datasets~\cite{LSMDCFIB,VideoQA} often rely on static or time-invariant information, which allows to find answers by observing a single frame with no consideration of temporal dependency.

To facilitate understanding of VideoQA models, we introduce a novel analysis framework, where we generate a customizable dataset using {\em Super Mario} video gameplays, referred to as {\em MarioQA}.
Our dataset is automatically generated from gameplay videos and question templates to contain desired properties for analysis. 
We employ the proposed framework to analyze the impact of a properly constructed dataset to answering questions, where we are particularly interested in the questions related to temporal reasoning of events.
The generated dataset consists of three subsets, each of which contains questions with a different level of difficulty in temporal reasoning.
Note that, by controlling complexity of questions, individual models can be trained and evaluated on different subsets.
Due to its synthetic nature, we can eliminate ambiguity in answers, which is often problematic in existing datasets, and make evaluation more reliable.

Our contribution is three-fold as summarized below:
\begin{itemize}
\item We propose a novel framework for analyzing VideoQA models, where a customized dataset is automatically generated to have desired properties for target analysis.
\item We generate a synthetic VideoQA dataset, referred to as MarioQA, using Super Mario gameplay videos with their logs and a set of predefined templates to understand temporal reasoning capability of models.
\item We present the benefit of our framework to facilitate analysis of algorithms and show that a properly generated dataset is critical to performance improvement.
\end{itemize}

The rest of the paper is organized as follows. 
We first review related work in Section~\ref{relwork}. 
Section~\ref{dataset} and \ref{spatiotemporal} present our analysis framework with a new dataset, and discuss several baseline neural models, respectively.
We analyze the models in Section~\ref{experiments}, and conclude our paper in Section~\ref{conclusion}.

%% file: related_work.tex

\begin{figure*}[t]
\centering
\includegraphics[width=0.98\linewidth]{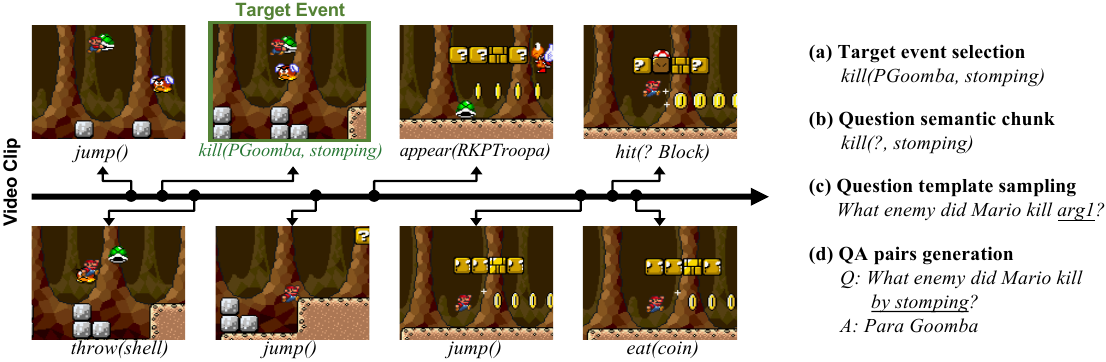}
\vspace{-0.3cm}
\caption{\small
Overall QA generation procedure. Given a gameplay video and event logs shown on the left, (a) target event is selected (marked as a green box), (b) question semantic chunk is generated from the target event, (c) question template is sampled from template pool, and (d) QA pairs are generated by filling the template and the linguistically realizing answer.
}
\label{fig:dataset_eg}
\vspace{-0.2cm}
\end{figure*}

\section{Related Work}
\label{relwork}

\subsection{Synthetic Testbeds for QA Models}
The proposed method provides a framework for VideoQA model analysis. 
Likewise, there have been several attempts to build synthetic testbeds for analyzing QA models~\cite{CLEVR,bAbI}.
For example, bAbI~\cite{bAbI} constructs a testbed for textual QA analysis with multiple synthetic subtasks, each of which focuses on a single aspect in textual QA problem.
The datasets are generated by simulating a virtual world given a set of actions by virtual actors and a set of constraints imposed on the actors.
For ImageQA, CLEVR~\cite{CLEVR} dataset is recently released to understand visual reasoning capability of ImageQA models.
It aims to analyze how well ImageQA models generalizes compositionality of languages.
Images in CLEVR are synthetically generated by randomly sampling and rendering multiple predetermined objects and their relationships.

\subsection{VideoQA Datasets}
Zhu \etal~\cite{VideoQA} constructed a VideoQA dataset in three domains using existing videos with grounded descriptions, where the domains include cooking scenarios~\cite{TACOS}, movie clips~\cite{MPII} and web videos~\cite{TRECVID}.
The fill-in-the-blank questions are automatically generated by omitting a phrase (a verb or noun phrase) from the grounded description and the omitted phrase becomes the answer.
For evaluation, the task is formed as answering multiple choice questions with four answer candidates. 
Similarly, \cite{LSMDCFIB} introduces another VideoQA dataset with automatically generated fill-in-the-blank questions from LSMDC movie description dataset.
Although the task for these datasets has a clear evaluation metric, the evaluation is still based on exact word matching rather than matching their semantics.
Moreover, the questions can be answered by simply observing any single frame without need for temporal reasoning.

MovieQA dataset~\cite{MovieQA} is another public benchmark of VideoQA based on movie clips.
This dataset contains additional information in other modalities including plot synopses, subtitles, DVS and scripts.
The question and answer (QA) pairs are manually annotated based on the plot synopses without watching movies.
The tasks in MovieQA dataset are difficult because the most questions are about the story of movies rather than about the visual contents of video clips. 
Hence, it needs to refer to external information other than the video clips, and not appropriate to evaluate trained models in terms of video understanding capability.

Contrary to these datasets, MarioQA is composed of videos with multiple events and event-centric questions.
Data in MarioQA require video understanding over multiple frames for reasoning temporal relationship between events but do not need extra information to find answers.

\subsection{Image and Video Question Answering}

\paragraph{ImageQA}
Because ImageQA needs to handle two input modalities, \ie, image and question, \cite{MCBPooling} presents a method of fusing two modalities to obtain rich multi-modal representations.
To handle the diversity of target tasks in ImageQA, \cite{composeVQA} and \cite{DPPNet} propose adaptive architecture design and parameter setting techniques, respectively.
Since questions often refer to particular objects within input images, many networks are designed to attend to relevant regions only~\cite{composeVQA,abcAtt,whereAtt,askAtt,stackedAtt}.
However, it is not straightforward to extend the attention models learned in ImageQA to VideoQA tasks due to additional temporal dimension in videos.

\vspace{-0.2cm}
\paragraph{VideoQA}
In MovieQA~\cite{MovieQA}, questions depend heavily on the textual information provided with movie clips, so models are interested in embedding the multi-modal inputs on a common space.
Video features are obtained by simply average-pooling image features of multiple frames.
On the other hand, \cite{VideoQA} employs gated recurrent units (GRU) for sequential modeling of videos instead of simple pooling methods.
In addition, unsupervised feature learning is performed to improve video representation power.
However, none of these methods explore attention models although visual attention turns out to be effective in ImageQA~\cite{composeVQA,abcAtt,whereAtt,askAtt,stackedAtt}.

%% file: dataset.tex

\section{VideoQA Analysis Framework}
\label{dataset}
This section describes our VideoQA analysis framework for temporal reasoning capability using MarioQA dataset.

\subsection{MarioQA Dataset Construction}
MarioQA is a new VideoQA dataset in which videos are recorded from gameplays and questions are about the events occurring in the videos.
We use the {\it Infinite Mario Bros.}\footnote{https://github.com/cflewis/Infinite-Mario-Bros} game, which is a variant of Super Mario Bros. with endless random level generation, to collect video clips with event logs and generate QA pairs automatically from extracted events using manually constructed templates.
Our dataset mainly contains questions about temporal relationships of multiple events to analyze temporal reasoning capability of models.
Each example consists of a $240\times 320$ video clip containing multiple events and a question with corresponding answer.
Figure~\ref{fig:dataset_eg} illustrates our data collection procedure.

\vspace{-0.3cm}
\paragraph{Design principle}
We build the dataset based on two design principles to overcome the existing limitations. 
First, the dataset is aimed to verify model’s temporal reasoning capability of events in videos. 
To focus on this main issue, we remove questions that require additional or external information to return correct answers and highlight model capacity for video understanding. 
Second, given a question, the answer should be clear and unique to ensure meaningful evaluation and interpretation. 
Uncertainty in answers, ambiguous linguistic structure of questions, and multiple correct answers may result in inconsistent or even wrong analysis, and make algorithms stuck in local optima. 

\vspace{-0.3cm}
\paragraph{Domain selection}
We choose gameplay videos as our video domain due to the following reasons.
First, we can easily obtain a large amount of videos that contain multiple events with their temporal dependency.
Second, learning complete semantics in gameplay videos is relatively easy compared to other domains due to their representation simplicity.
Third, occurrence of an event is clear and there is no perceptual ambiguity.
In real videos, answers for a question may be diverse depending on annotators because of subjective perception of visual information. 
On the contrary, we can simply access the oracle within the code to find answers.

\begin{table*}
\centering
\caption{\small Comparisons of the three VideoQA datasets.} \vspace{-0.3cm}
\label{tab:datasets}
\scalebox{0.78}{
\begin{tabular}{
m{2.95cm}|@{}C{1.1cm}@{}|@{}C{1.1cm}@{}|m{5.2cm}|m{3cm}|m{2.4cm}|m{2.9cm}|@{}C{1.4cm}@{}
}
Dataset					& Video 		& Extra Info.	& Goal							& Domain											& Data Source												& Method				& \# of QA	 \\ \hline
Fill-in-the-blank~\cite{VideoQA}	& \checkmark	& --			& filling blanks of video descriptions		& \parbox[c]{3cm}{cooking scenario, \\movies, web videos} 	& \parbox[c][0.9cm][c]{2.5cm}{video description \\dataset}						& automatic generation	& 390,744	 \\
MovieQA~\cite{MovieQA}		& \checkmark	& \checkmark	& answering questions for movie stories	& movies											& \parbox[c][1.4cm][c]{2.5cm}{plot synopses, \\movies, subtitles, \\DVS, scripts}	& human annotation		& 15,000	 \\
MarioQA (ours)				& \checkmark	& --			& answering event-centric questions by temporal reasoning	& \parbox[c]{3cm}{gameplay \\(Infinite Mario Bros.)}			& \parbox[c][0.9cm][c]{2.5cm}{gameplays, \\QA templates}				& automatic generation	& 187,757	 \\
\end{tabular}
}
\end{table*}

\begin{table*}
\centering
\caption{\small Examples of QA pairs in the three VideoQA datasets with answers boldfaced.} 
\label{tab:videoqa_examples}
\vspace{-0.1cm}
\scalebox{0.8}{
\begin{tabular}{
m{6.8cm}|m{6.8cm}|m{6.8cm}
}
Fill-in-the-blank~\cite{VideoQA}	& MovieQA~\cite{MovieQA}	 & MarioQA (ours) 	 \\ \hline
\parbox[c]{6.8cm}{
\begin{itemize}[noitemsep,leftmargin=*,nolistsep]
\item The man grab a \_\_\_\_\_\_.~~/~~{\bf plate}
\item He slice the  \_\_\_\_\_\_ very thinly, run through slice them twice.~~/~~{\bf garlic}
\item The woman wash the plum off with \_\_\_\_\_\_ and the towel. ~~/~~{\bf water}
\item People eat and \_\_\_\_\_\_ at a resort bar~~/~~{\bf drink}
\item A little \_\_\_\_\_\_ play hide and seek with a woman.~~/~~{\bf girl}
\item a small group of guy climb \_\_\_\_\_\_ and boulder in a forest~~/~~{\bf rock}
\end{itemize}
} &
\parbox[c]{6.8cm}{
\begin{itemize}[noitemsep,leftmargin=*,nolistsep]
\item What is Rob's passion in life?~~/~~{\bf Music}
\item How many years does Lucas serve in prison?~~/~~{\bf 15 years}
\item How does Epps's wife demonstrate her attitude to Patsey?~~/~~{\bf By constantly humiliating and degrading her}
\item Who eventually hires Minny?~~/~~{\bf Celia Foote}
\item When was Milk assassinated?~~/~~{\bf 1978}
\item Does Solomon get the promised job?~~/~~{\bf No}
\end{itemize}
} &
\parbox[c][4.3cm][c]{6.8cm}{
\begin{itemize}[noitemsep,leftmargin=*,nolistsep]
\item Where was the Green Koopa Troopa stomped?~~/~~{\bf Hill}
\item Which enemy was killed by Mario's stomp after Mario hit a coin block?~~/~~{\bf Goomba}
\item How many fireballs did Mario shoot? ~~/~~{\bf 5}
\item How many times did Mario jump before holding a shell?~~/~~{\bf 3}
\item What is the type of stage?~~/~~{\bf Cave}
\item What is the state of Mario when eating a mushroom?~~/~~{\bf Fire form}
\end{itemize}
}
\end{tabular}
}
\vspace{-0.1cm}
\end{table*}

\vspace{-0.2cm}
\paragraph{Target event and clip extraction}
We extract 11 distinct events $E=\text{\{kill, }$die, jump, hit, break, appear, shoot, throw, kick, hold, eat\} with their arguments, \eg, {\it agent}, {\it patient} and {\it instrument} from gameplays.
For each extracted event as a target, we randomly sample video clips containing the target event with duration of 3 to 6 seconds.
We then check the uniqueness of the target event within the sampled clip. 
For instance, the event {\it kill} with its arguments {\it PGoomba} and {\it stomping} is a unique target event among 8 extracted events in the video clip of Figure~\ref{fig:dataset_eg}.
This uniqueness check process rejects questions involving multiple answers since they cause ambiguity in evaluation.

\vspace{-0.2cm}
\paragraph{Template-based question and answer generation}
Once the video clips of unique events are extracted, we generate QA pairs from the extracted events.
We randomly eliminate one of the event arguments to form a {\it question semantic chunk} and generate a question from the question semantic chunk using predefined question templates.
For example, an argument {\it PGoomba} is removed to form a question semantic chunk, {\it kill(?,~stomping)}, from the event {\it kill(PGoomba, stomping)} in Figure~\ref{fig:dataset_eg}.
Then, a question template {\it `What enemy did Mario kill \underline{~arg1~}?'} is selected from the template pool for the question semantic chunk. 
Finally, a question is generated by filling the template with a phrase {\it `by~stomping'}, which linguistically realizes an argument, {\it stomping}.
We use the template-based question generation because it allows to control level of semantics required for question answering.
When annotators are told to freely create questions, it is hard to control the required level of semantics.
So, we ask human annotators to create multiple linguistic realizations of a question semantic chunk.
After question generation, a corresponding answer is also generated from the eliminated event argument, \ie, {\it Para Goomba} in the above example.
Note that the dataset is easily customized by updating QA templates to further reflect analytical perspectives and demands.

\begin{figure*}
\centering
\begin{subfigure}[t]{0.3\linewidth}
\centering
\includegraphics[width=1\linewidth]{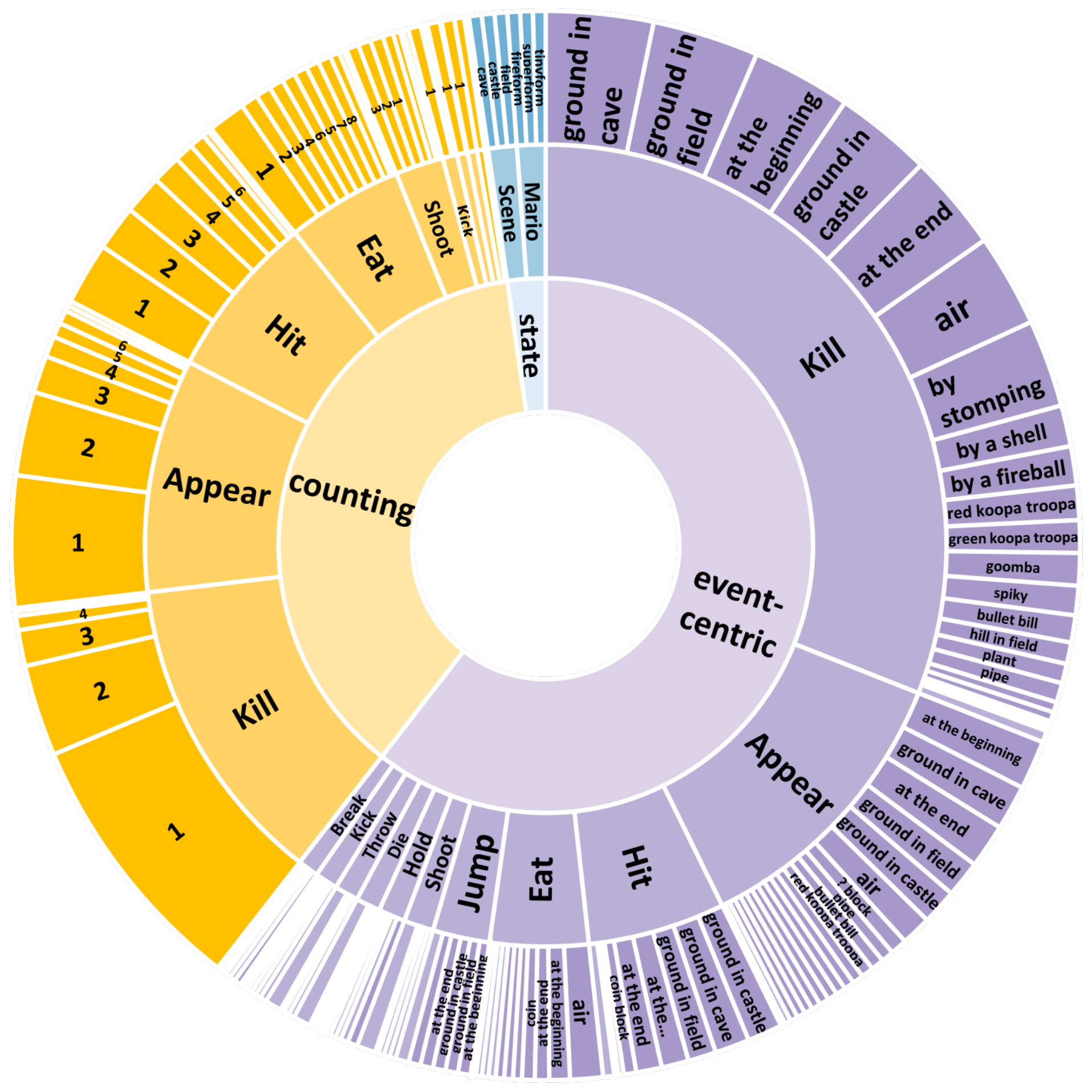}
\vspace{-0.5cm}
\subcaption{NT}
\end{subfigure}
~~~~
\begin{subfigure}[t]{0.3\linewidth}
\centering
\includegraphics[width=1\linewidth]{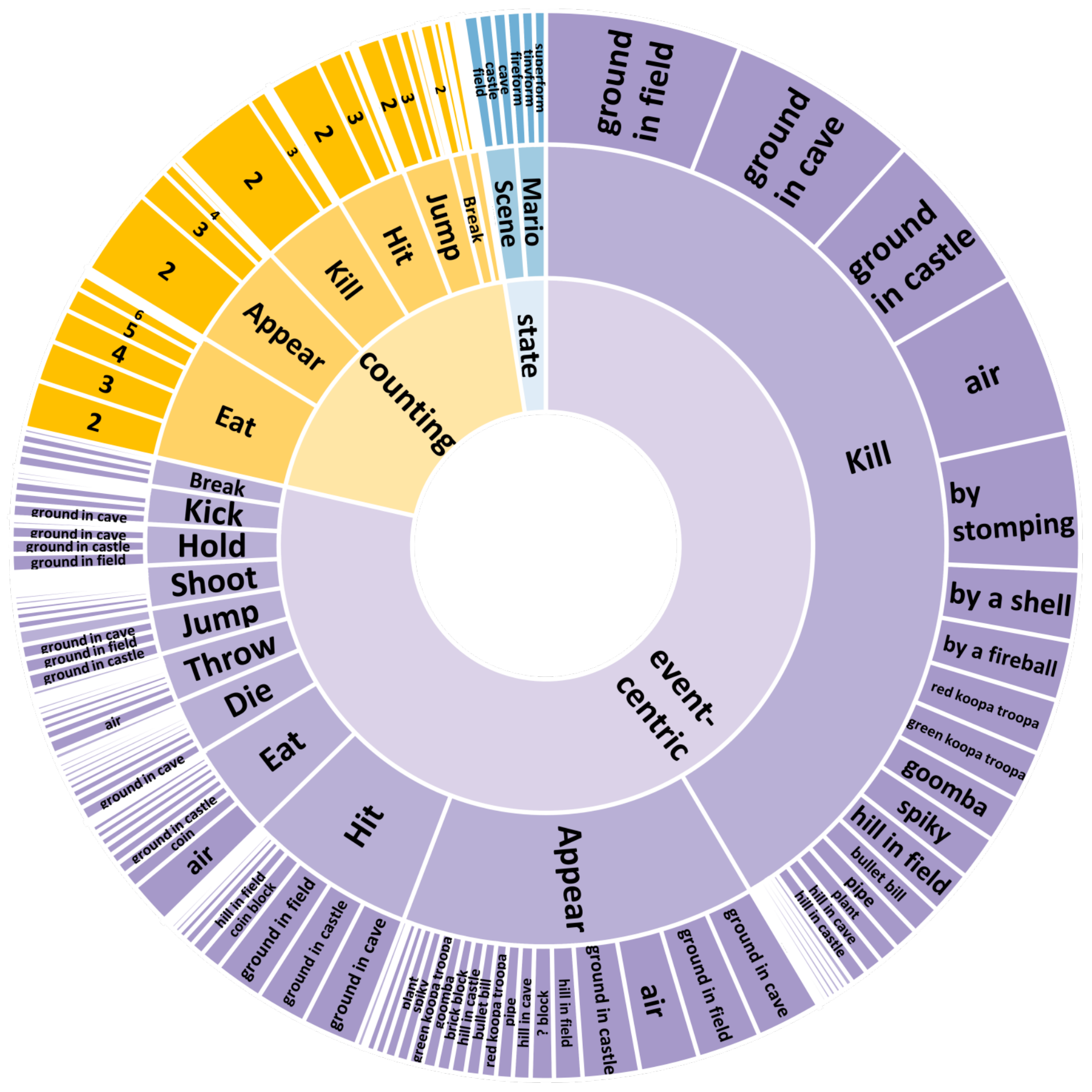}
\vspace{-0.5cm}
\subcaption{ET}
\end{subfigure}
~~~~
\begin{subfigure}[t]{0.3\linewidth}
\centering
\includegraphics[width=1\linewidth]{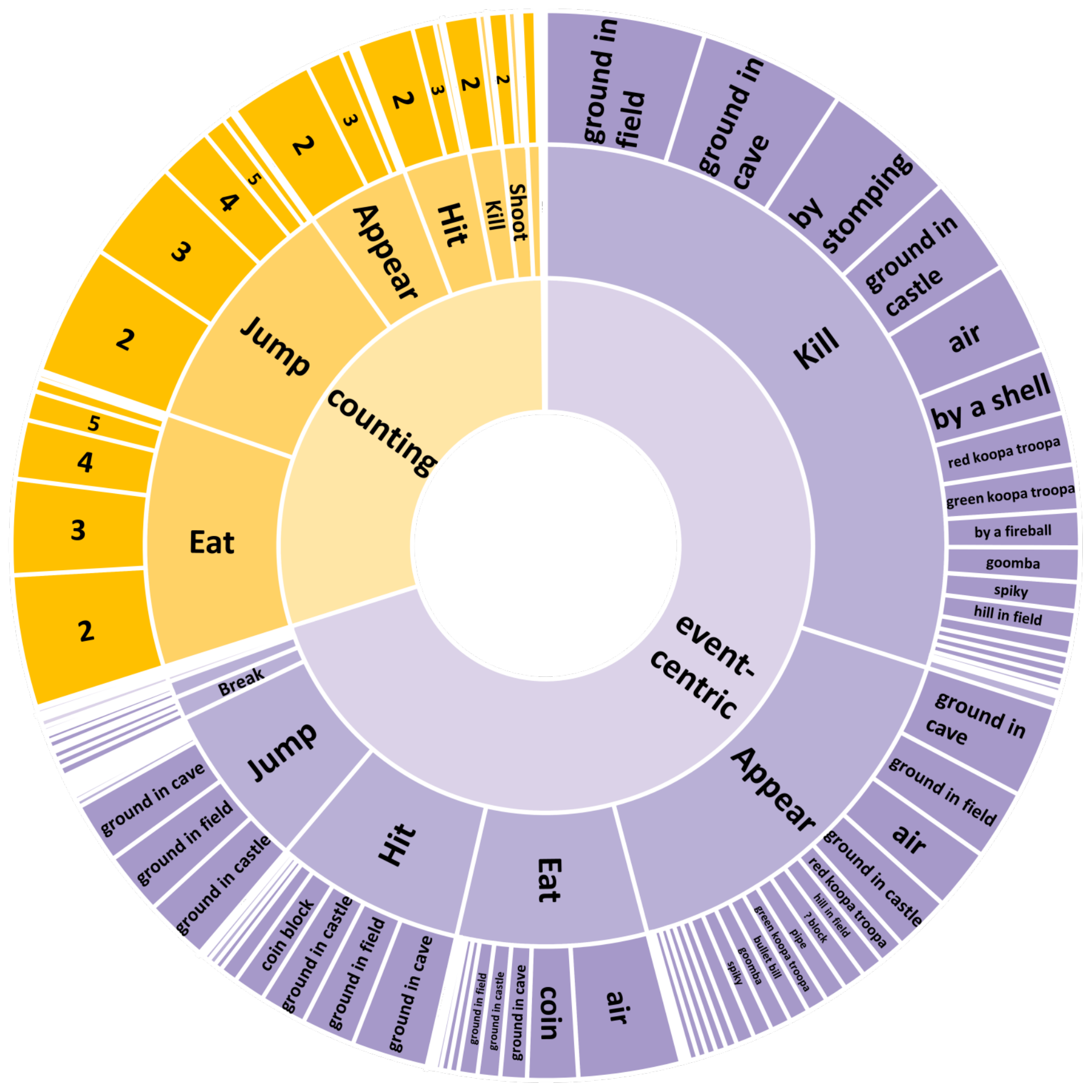}
\vspace{-0.5cm}
\subcaption{HT}
\end{subfigure}
\vspace{-0.2cm}
\caption{\small The distributions of question and answer pairs in three subsets of MarioQA. 
}
\label{fig:question_dist}
\end{figure*}

\subsection{Characteristics of Dataset for Model Analysis}
\label{sub:characteristics}
There are three question types in MarioQA to maintain diversity.
The followings are examples of event-centric, counting and state questions, respectively: {\it `What did Mario hit before killing Goomba?'}, {\it `How many coins did Mario eat after a Red Koopa Paratroopa appears?'} and {\it `What was Mario's state when Green Koopa Paratroopa appeared?'}
While questions in the three types generally require observation over multiple frames to find answers, a majority of state questions just need a single frame observation about objects and/or scene due to the uniqueness of the state within a clip.

As seen in the above examples, multiple events in a single video may be temporally related to each other, and understanding such temporal dependency is an important aspect in VideoQA.
In spite of importance of this temporal dependency issue in videos, it has not been explored explicitly due to lack of proper datasets and complexity of tasks.
Thanks to the synthetic property, we can generate questions about temporal relationships conveniently.

We construct MarioQA dataset with three subsets, which contain questions with different characteristics in temporal relationships: questions with no temporal relationship (NT), with easy temporal relationship (ET) and with hard temporal relationships (HT).
NT asks questions about unique events in the entire video without any temporal relationship phrase.
ET and HT have questions with temporal relationships in different levels of difficulty.
While ET contains questions about globally unique events, HT involves distracting events making a VQA system choose a right answer out of multiple identical events using temporal reasoning;
for a target event {\it kill(PGoomba, stomping)}, any {\it kill(*, *)} events in the same video clip are considered as distracting events.
Note that the answer of a question {\it 'How many times did Mario jump after throwing a shell?'} about the video clip in Figure~\ref{fig:dataset_eg} is not 3 but 2 due to its temporal constraint.
Note that the generated questions are still categorized into one of three types---event-centric, counting and state questions.

\subsection{Dataset Statistics}

From a total of 13 hours of gameplays, we collect 187,757 examples with automatically generated QA pairs.
There are 92,874 unique QA pairs and each video clip contains 11.3 events in average.
There are 78,297, 64,619 and 44,841 examples in NT, ET and HT, respectively.
Note that there are 3.5K examples that can be answered using a single frame of video; the portion of such examples is only less than 2\%. 
The other examples are event-centric; 98K examples require to focus on a single event out of multiple ones while 86K need to recognize multiple events for counting (55K) or identifying their temporal relationships (44K).
Note that there are instances that belong to both cases.

Some types of events are more frequently observed than others due to the characteristics of the game, which is also common in real datasets.
To make our dataset more balanced, we have a limit for the maximum number of same QA pairs. 
The QA pair distribution of each subset is depicted in Figure~\ref{fig:question_dist}.
The innermost circles show the distributions of the three question types.
The portion of event-centric questions is much larger than those of the other types in all three subsets as we focus on the event-centric questions. 
The middle circles present instance distributions in each question type, where we observe a large portion of {\it kill} event since {\it kill} events occur with more diverse arguments such as multiple kinds of enemies and weapons.
The outermost circles show the answer distributions related to individual events or states.

The characteristics of VideoQA datasets are presented in Table~\ref{tab:datasets}.
The number of examples in \cite{VideoQA} is larger than the other datasets but fragmented into many subsets, which are hard to be used as a whole.
MovieQA~\cite{MovieQA} dataset has extra information in other modalities.
Both datasets have limitations in evaluating model capacity for video understanding as in the examples in Table~\ref{tab:videoqa_examples}.
The questions in \cite{VideoQA} are mainly about the salient contents throughout the videos.
These questions can be answered by understanding a single frame rather than multiple ones.
On the other hand, the questions in MovieQA are often too difficult to answer by watching videos as they require very high-level abstraction about movie story.
In contrast, MarioQA contains videos with multiple events and their temporal relationships.
The event-centric questions with temporal dependency allow us to evaluate whether the model can reason temporal relationships between multiple events.

%% file: spatiotemporal.tex

\begin{figure*}[t]
\centering
\includegraphics[width=1\linewidth]{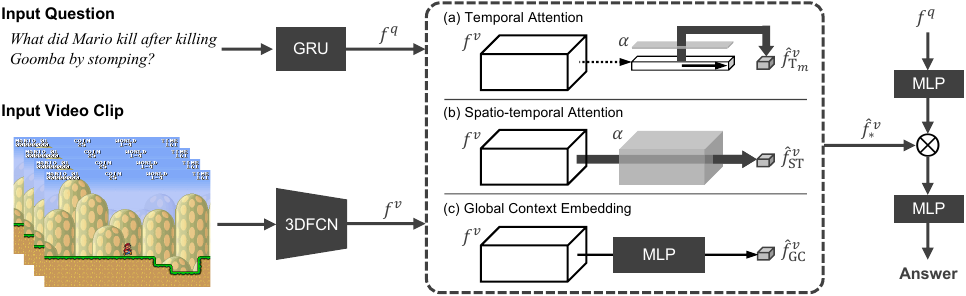}
\vspace{-0.3cm}
\caption{\small
Architecture of the neural models in our analysis. 
The encoders---3DFCN and GRU for video feature extraction and question embedding, respectively---as well as the final classification network are shared by all neural models.
We introduce neural baseline models with three different attention mechanisms on MarioQA: (a) temporal attention, (b) spatio-temporal attention and (c) global context embedding with no attention.
}
\label{fig:video_net}
\end{figure*}

\section{Neural Models for MarioQA}
\label{spatiotemporal}

We describe our neural baseline models for MarioQA.
All networks comprise of three components: question embedding, video embedding and classification networks depicted in Figure~\ref{fig:video_net}.
We explore each component in detail below.

\subsection{Question Embedding Network}
Recurrent neural networks (RNN) with memory units such as long short term memory (LSTM) or gated recurrent unit (GRU) are widely used in ImageQA~\cite{VQA,MCBPooling,DPPNet,askMe,dynamicMem}.
Following~\cite{DPPNet} and \cite{VideoQA}, we obtain a question embedding vector using the pretrained GRU on the large corpus~\cite{skipthought}. 
The question embedding vector $f^q \in \mathbb{R}^{2400}$ is given by
\begin{align}
f^q&=\mathrm{SkipThought}(q)
\end{align}
where $\mathrm{SkipThought}(\cdot)$ denotes a pretrained question embedding network and $q$ is an input question.

\subsection{Video Embedding Network}
We employ a 3D fully convolutional network (3DFCN) for video feature extraction.
The 3DFCN is composed of five $3\times3\times3$ convolution layers with four pooling layers between them.
The first two pooling layers only pool features in the spatial dimensions but not in the temporal dimension whereas the last two pooling layers pool in all spatio-temporal dimensions.
We rescale input videos to $120\times160$ and sample $K$ key frames with temporal stride 4 before feeding to the network.
The output of the 3DFCN is a feature volume $\bm{f}^v \in \mathbb{R}^{512\times T\times 7\times 10}$, which is given by
\begin{equation}
\bm{f}^v = \mathrm{3DFCN}(V),
\end{equation}
where $T = K/4$ and $V$ denotes an input video.
Once the video features are extracted from the 3DFCN, we embed these features volume $\bm{f}^v$ onto a low-dimensional spaces using one of the following ways.

\subsubsection{Embedding with Temporal Attention}
As many events in a video from MarioQA are not relevant to input question, temporal localization of target events in videos may be important to answer questions.
The temporal attention model embeds the entire video feature volume onto a low-dimensional space by interpolating features in the temporal dimension based on the weights learned for attention.
In this model, a frame feature is first extracted through an average pooling from the feature map at every location in the temporal dimension.
Formally, a frame feature $f^\mathrm{frame}_t$ at time $t$ is obtained from a video feature map $\bm{f}^v_t$ by
\begin{equation}
f^\mathrm{frame}_t = \underset{i, j}{\mathrm{avgpool}}(\bm{f}^v_t),
\end{equation}
where ${\mathrm{avgpool}}_{i,j}$ denotes the average pooling operation in spatial dimensions and the resulting frame features are used for temporal attention process.

%
\vspace{-0.2cm}
\paragraph{Single-step temporal attention}
Given a frame feature $f^\mathrm{frame}_t$ and a question embedding $f^q$, the temporal attention probability $\alpha_{t}^{1}$ for each frame feature is obtained by
\begin{align}
s_{t}^{1}&=\mathrm{att}(f^\mathrm{frame}_{t},f^q), \\
\alpha_t^1 &= \text{softmax}_t (\bm{s}^1)
\end{align}
where $\mathrm{att}(\cdot,\cdot)$ is a multilayer perceptron composed of one hidden layer with 512 nodes and a scalar output.
The temporally attended video embedding $\hat{f}^v_{\mathrm{T}_1}$ is obtained using the attention probabilities as
\begin{equation}
\hat{f}^v_{\mathrm{T}_1} = \sum_t^T \alpha_{t}^{1}f^\mathrm{frame}_{t}.
\label{eq:temporally_attended_embedding_1}
\end{equation}

\vspace{-0.2cm}
\paragraph{Multi-step temporal attention}
We also employ multi-step temporal attention models, which applies the temporal attention multiple times.
We follow the multiple attention process introduced in~\cite{stackedAtt}.
The temporally attended video embedding with $m$ steps denoted by $\hat{f}^v_{\mathrm{T}_m}$ is obtained by the same process except that the question embedding is refined by adding the previous attended embedding $\hat{f}^v_{\mathrm{T}_{m-1}}$ as
\begin{align}
s_{t}^{m}&=\mathrm{att}(f^\mathrm{frame}_{t},f^q+\hat{f}^v_{\mathrm{T}_{m-1}}), \\
\alpha_t^m &= \text{softmax}_t (\bm{s}^m),
\end{align}
where $\hat{f}^v_{\mathrm{T}_0}$ is a zero vector.
The temporally attended video embedding $\hat{f}^v_{\mathrm{T}_m}$ with $m$ steps is given by a similar linear combination in Eq.~\eqref{eq:temporally_attended_embedding_1}.
%
%
%
\subsubsection{Embedding with Spatio-Temporal Attention}
We can attend to a single feature in a spatio-temporal feature volume.
The attention score $s_{t,i,j}$ for each feature $f^v_{t,i,j}$ and the attention probability $\alpha_{t,i,j}$ is given respectively by
\begin{align}
s_{t,i,j} &= \mathrm{att}(f^v_{t,i,j},f^q) ~~~\text{and} \\
\alpha_{t,i,j} &= \mathrm{softmax}_{t,i,j}(\bm{s}),
\end{align}
where the softmax function is applied throughout the spatio-temporal space to normalize the attention probabilities.
The spatio-temporally attended video embedding $\hat{f}^v_\mathrm{ST}$ is then obtained by
\begin{equation}
\hat{f}^v_\mathrm{ST} = \sum_t\sum_i\sum_j \alpha_{t,i,j}f^v_{t,i,j}.
\end{equation}

\begin{table*}
\centering
\caption{\small
Accuracies [\%] for the models on test splits. Models are trained on different combinations of subsets: NT, NT+ET and NT+ET+HT to see the impact of each subset on accuracies. 
Each trained model is then tested on test split of each subset.
} 
\vspace{-0.1cm}
\label{tab:res_types}
\vspace{-0.2cm}
\scalebox{0.8}{
\begin{tabular}{
@{}C{2cm}@{}|@{}C{1.6cm}@{}@{}C{1.6cm}@{}@{}C{1.6cm}@{}@{}C{1.6cm}@{}|@{}C{1.6cm}@{}@{}C{1.6cm}@{}@{}C{1.6cm}@{}@{}C{1.6cm}@{}|@{}C{1.6cm}@{}@{}C{1.6cm}@{}@{}C{1.6cm}@{}@{}C{1.6cm}@{}
}
	& \multicolumn{4}{c|}{Trained on NT (case 1)} 		& \multicolumn{4}{c|}{Trained on NT+ET (case 2)}	& \multicolumn{4}{c}{Trained on NT+ET+HT (case 3)} 	\\
	& NT		& ET		& HT		& ALL	& NT		& ET		& HT		& ALL	& NT		& ET		& HT		& ALL		\\ \hline
V	& 21.29	& 32.33	& 31.36	& 27.49	& 20.67	& 35.80	& 34.25	& 29.12	& 21.16	& 35.32	& 34.00	& 29.10		\\
Q	& 40.86	& 24.65	& 26.50	& 31.85	& 40.23	& 35.52	& 36.46	& 37.71	& 39.79	& 35.67	& 39.65	& 38.34		\\ 
AP	& 56.76	& 33.14	& 29.39	& 42.09	& 58.79	& 62.57	& 59.02	& 60.15	& 60.03	& 66.49	& 64.95	& 63.43		\\ \hline
1-T	& 56.65	& 42.99	& 41.34	& 48.29	& 61.61	& 64.21	& 60.20	& 62.17	& 64.28	& 69.64	& 67.21	& 66.82		\\
2-T 	& 53.96	& 39.32	& 40.71	& 45.76	& 62.87	& 66.42	& 61.75	& 63.82	& 64.05	& 66.13	& 65.65	& 65.15		\\ \hline
ST	& 60.18	& 47.79	& 45.87	& 52.50	& 62.80	& 68.44	& 61.87	& 64.52	& 66.38	& 72.73	& 69.27	& 69.26		\\
GC	& 55.83	& 29.89	& 25.24	& 39.60	& 65.62	& 74.27	& 61.35	& 67.58	& 66.47	& 75.10	& 68.89	& 70.02		\\
\end{tabular}
}
\vspace{-0.3cm}
\end{table*}

\subsubsection{Global Context Embedding}
A popular video embedding method is to embed the entire video feature volume using fully-connected layers~\cite{vidFeat,C3D}.
This method is more appropriate to capture the global context than the attention-based video embedding models described above.
We build a multi-layer perceptron that has two fully-connected layers with 512 hidden nodes and 512 dimensional output layer for video embedding.
That is, the video embedding $\hat{f}^v_\mathrm{GC}$ is obtained by
\begin{equation}
\hat{f}^v_\mathrm{GC} = \mathrm{vemb}(\bm{f}^v_\mathrm{flat})
\label{eq:gce}
\end{equation}
where $\mathrm{vemb}$ is the MLP for the video embedding and $\bm{f}^v_\mathrm{flat}$ is the flattened video feature volume.

%
\subsection{Classification Network}
The classification network predicts the final answer given a video embedding $\hat{f}^v_*$ and a question embedding $f^q$, where $* \in \{ \mathrm{T}_1, \mathrm{T}_m, \mathrm{ST}, \mathrm{GC} \}$.
The question embedding $f^q$ is first transformed to the common space with the video embedding $\hat{f}^v$, and the transformed embedding is given by
\begin{equation}
\hat{f}^q = \sigma(W_qf^q),
\end{equation}
where $W_q$ is $512\times2400$ weight matrix and $\sigma$ is a nonlinear function such as ReLU.
Then, we fuse the two embedded vectors and generate the final classification score by
\begin{equation}
S = \mathrm{softmax}(W_\mathrm{cls}(\hat{f}^v \odot \hat{f}^q)),
\end{equation}
where $W_\mathrm{cls}$ is the weight matrix for final classification layer.

%% file: experiments.tex

\begin{table}
\centering
\caption{\small Number of examples in each split of subsets.
The training, validation and test splits in each subset are obtained by random sampling of 60\%, 20\%, 20\% of data from each subset.}
\label{tab:splits}
\vspace{-0.2cm}
\scalebox{0.8}{
\begin{tabular}{
@{}C{1.5cm}@{}|@{}C{1.75cm}@{}@{}C{1.75cm}@{}@{}C{1.75cm}@{}|@{}C{1.75cm}@{}
}
	& train	& valid	& test	& total	\\ \hline
NT	& 46,978	& 15,660	& 15,659	& 78,297	\\
ET	& 38,771	& 12,924	& 12,924	& 64,619	\\
HT	& 26,904	& 8,969	& 8,968	& 44,841	\\
\end{tabular}
}
\vspace{-0.3cm}
\end{table}

\begin{figure*}
\centering
\includegraphics[width=0.93\linewidth]{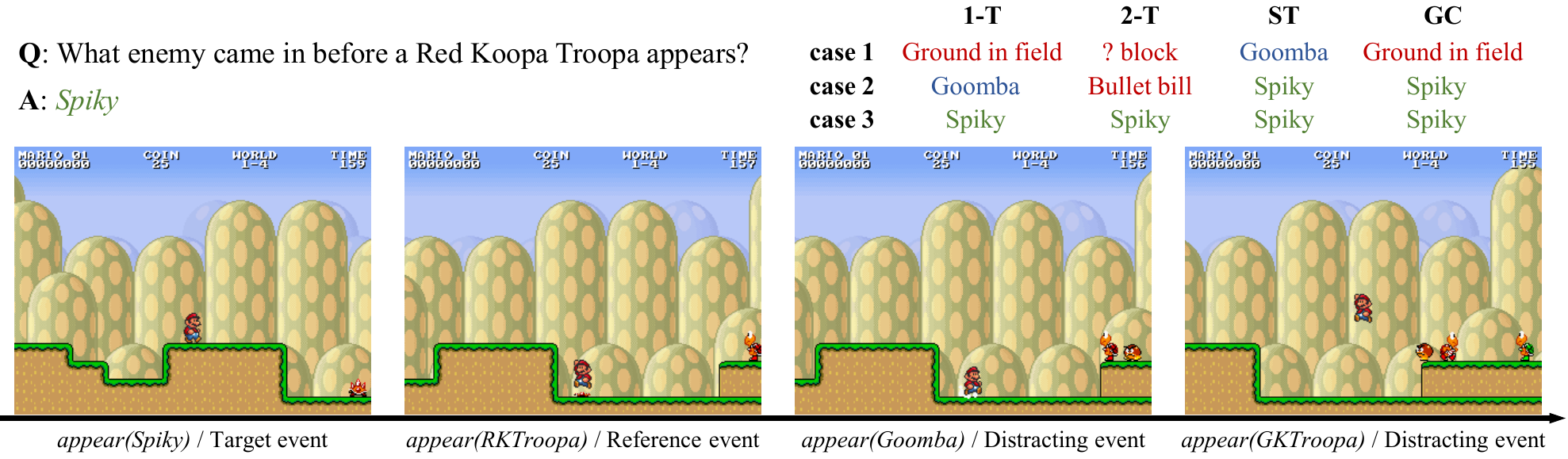}
\vspace{-0.2cm}
\caption{\small
Qualitative results of an HT question. 
Four frames of a video clip representing target, reference and two distracting events are presented with their events, where reference event means temporally related event to target event in question.
Models are trained on different combinations of subsets: NT (case~1), NT+ET (case~2) and NT+ET+HT (case~3). 
Note that models tend to predict correct answers when trained on properly constructed training dataset while most models generate answers far from correct ones in case~1.
Correct and incorrect answers are marked as green and red respectively while incorrect answers from distracting events are marked as blue.
}
\vspace{-0.4cm}
\label{fig:qualitative}
\end{figure*}

\section{Experiments}
\label{experiments}
\subsection{Experimental Setting}

We have three subsets in MarioQA dataset as presented in Table~\ref{tab:splits}. 
We aim to analyze the impact of questions with temporal relationships in training, so we train models on the following three combinations of the subsets: NT (case~1), NT+ET (case~2) and NT+ET+HT (case~3).
Then, these models are evaluated to verify temporal reasoning capability on the test split of each subset. 
We implement two versions of the temporal attention models with one and two attention steps (1-T and 2-T) following \cite{stackedAtt}.
The spatio-temporal attention model (ST) and the global context embedding (GC) are also implemented.
In addition to these models, we build three simple baselines:
\begin{itemize}
\item{\bf Video Only (V)}~~Given a video, the model predicts an answer without knowing the question.
We perform video embedding by Eq.~\eqref{eq:gce} and predict answers using a multi-layer perceptron with one hidden layer.
\item{\bf Question Only (Q)}~~This model predicts an answer by observing questions but without seeing videos.
The same question embedding network is used with the classification.
\item{\bf Average Pooling (AP)}~~This model embeds the video feature volume $\bm{f}^v$ by average pooling throughout the spatio-temporal space and use it for final classification.
This model is for comparisons with the attention models as the average pooling is equivalent to assigning the uniform attention to every spatio-temporal location.
\end{itemize}
%
All the models are trained end-to-end by the standard backpropagation from scratch while the question embedding network is initialized with a pretrained model~\cite{skipthought}.
The vocabulary sizes of questions are 136, 168 and 168 for NT, ET and HT, respectively, and the number of answer classes is 57.

In our scenario, the initial model of each algorithm is trained with NT only and we evaluate algorithms in all three subsets by simply computing the ratio of correct answers to the total number of questions.
Then, we add ET and HT to training data one by one, and perform the same evaluation and observe tendency of performance change.

\begin{figure}
\centering
\includegraphics[width=1\linewidth]{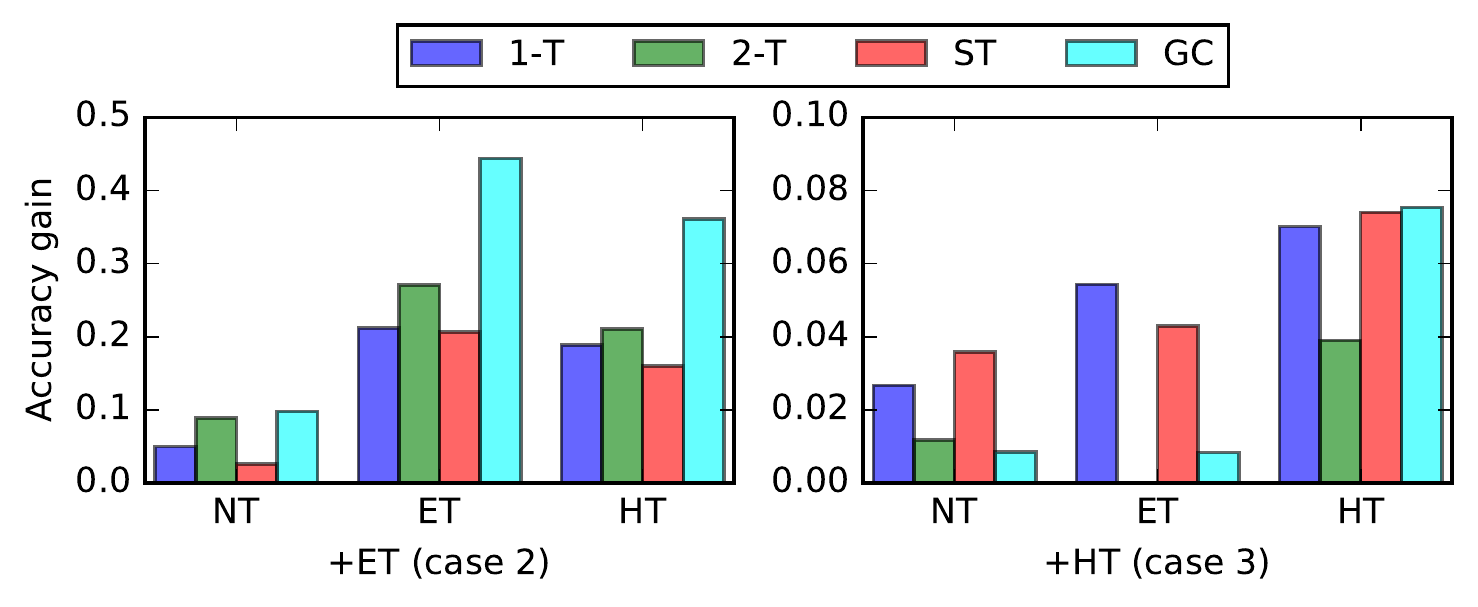}
\vspace{-0.4cm}
\caption{\small
Accuracy gain of every model on each subset whenever a subset is added to training set. 
Added subsets are shown below each graph and each bar in graphs shows gain from accuracy without added subset. 
Test subsets are shown on x-axis.
}
\label{fig:graph}
\end{figure}

\subsection{Results}

\paragraph{Analysis on neural models}
Table~\ref{tab:res_types} presents the overall results of our experiments.
Obviously, two simple baselines (V and Q) show significantly lower performance than the others\footnote{To demonstrate the strength of trained models, we evaluate random guess accuracies, which are 16.96, 12.66 and 12.66 (\%) for NT, ET and HT, respectively.
Also, The accuracies obtained by selecting the most frequent answer are 37.85, 32.29 and 39.00 (\%) for NT, ET and HT, respectively.}.
Although three attention-based models and AP outperform GC in case~1, GC becomes very competitive in case~2 and case~3.
It is probably because network architectures with general capabilities such as fully-connected layers are more powerful than the linear combinations of attentive features; if the dataset is properly constructed involving examples with temporal relationships, GC is likely to achieve high performance.
However, attention models may be able to gain more benefit from pretrained models, and GC is a more preferable model for our environment with short video clips.

\begin{table}
\centering
\caption{\small
Results from GC in the three subsets with roughly the same number of training examples.
} 
\label{tab:res_vq}
\vspace{-0.2cm}
\scalebox{0.8}{
\begin{tabular}{
@{}C{3.3cm}@{}|@{}C{2.7cm}@{}|@{}C{1.1cm}@{}@{}C{1.1cm}@{}@{}C{1.1cm}@{}@{}C{1.1cm}@{}
}
Train set			& \# of train example	& NT		& ET		& HT		& ALL	\\ \hline
NT (case 1)		& 46,978	& 55.83	& 29.89	& 25.24	& 39.60	\\
NT+ET (case 2)	& 46,990	& 53.39	& 58.54	& 60.16	& 56.78	\\
NT+ET+HT (case 3)	& 46,970	& 54.49	& 60.83	& 61.50	& 58.35	\\
\end{tabular}
}
\vspace{-0.3cm}
\end{table}

\vspace{-0.2cm}
\paragraph{Analysis on dataset variation}
Our results strongly suggest that proper training data construction would help to learn a better model.
Figure~\ref{fig:qualitative} presents qualitative results for an HT question in all three cases. 
It shows that the models tend to predict the correct answer better as ET and HT are added to training dataset.
The quantitative impact of adding ET and HT to training data is illustrated in Figure~\ref{fig:graph}.
By adding ET to training dataset, we observe improvement of all algorithms with attention models (and AP) in all three subsets, where performance gains in ET are most significant consistently in all algorithms.
The similar observation is found when HT is additionally included in the training dataset although the magnitudes of improvement are relatively small.
This makes sense because the accuracies are getting more saturated as more data are used for training. 

It is noticeable that training with the subsets that require more difficult temporal reasoning also improves performance of the subsets with easier temporal reasoning; training with ET or HT improves performance not only on ET and HT but also on NT.
It is also interesting that training with ET still improves the accuracy on HT.
Since ET does not contain any distracting events, questions in ET can be answered conceptually regardless of temporal relationships. 
However, the improvement on HT in case 2 intimates that the networks still learn a way of temporally relating events using ET.

One may argue that the improvement mainly comes from the increased number of training examples in case~2 and case~3. 
To clarify this issue, we train GC models for case~2 and case~3 using roughly the same number of training examples with case~1 (Table~\ref{tab:res_vq}).
Due to smaller training datasets, the overall accuracies are not as good as our previous experiment but the performance improvement tendency is almost same.
It is interesting that three cases on NT testing set achieve almost same accuracy in this experiment even with less training examples in NT for case~2 and case~3.
This fact shows that ET and HT are helpful to solve questions in NT.

%% file: conclusion.tex

\section{Conclusion}
\label{conclusion}
We propose a new analysis framework for VideoQA and construct a customizable synthetic dataset, MarioQA.
Unlike existing datasets, MarioQA focuses on event-centric questions with temporal relationships to evaluate temporal reasoning capability of algorithms.
The questions and answers in MarioQA are automatically generated based on manually constructed question templates.
We use our dataset for the analysis on the impact of questions with temporal relationships in training and show that properly collected dataset is critical to improve quality of VideoQA models.
However, we believe that MarioQA can be used for further analyses on other perspectives of VideoQA by customizing the dataset with preferred characteristics in its generation.